\documentclass[10pt,twocolumn,letterpaper]{article}

\usepackage[pagenumbers]{wacv}

\definecolor{wacvblue}{rgb}{0.21,0.49,0.74}
\usepackage[pagebackref,breaklinks,colorlinks,allcolors=wacvblue]{hyperref}

\def\wacvPaperID{1948}
\def\confName{WACV}
\def\confYear{2026}

\title{A LoRA is Worth a Thousand Pictures}

\author{
Chenxi Liu$^{1,*}$ \hspace{10pt}
Towaki Takikawa$^{2}$ \hspace{10pt}
Alec Jacobson$^{1}$ \\
{\small $^{1}$University of Toronto \hspace{20pt} $^{2}$Outerport} \\
}

\usepackage[mathletters]{ucs}
\usepackage{soul}
\usepackage[utf8]{inputenc}

\usepackage[ruled]{algorithm2e} %

\usepackage{boldline} %
\usepackage{multirow}
\usepackage{stfloats}
\usepackage{float}
\usepackage{bm}

\providecommand{\A}{}
\providecommand{\B}{}

\providecommand{\Q}{}

\providecommand{\c}{}

\providecommand{\l}{}

\providecommand{\q}{}

\providecommand{\s}{}
\providecommand{\t}{}

\providecommand{\x}{}

\usepackage{tabularx}
\usepackage{booktabs}
\usepackage{makecell}
\definecolor{white}{rgb}{1,1,1}
\definecolor{lightbluishgrey}{rgb}{0.76471,0.84824,0.91647}

\usepackage{subcaption}
\usepackage{wrapfig}
\usepackage{graphicx}

\usepackage{listings}
\usepackage{layouts}
\makeatletter 
\newcommand{\layoutdetails}{%
\begin{tabular}{ll}
 \texttt{\textbackslash{textwidth}} & \printinunitsof{in}\prntlen{\textwidth} \\
\texttt{\textbackslash{linewidth}} & \printinunitsof{in}\prntlen{\linewidth} \\
Main text font &  \f@size pt \f@family \\
\sffamily \small Caption text font &  \sffamily \small \f@size pt \f@family \\
\end{tabular}%
}
\makeatother

\renewcommand{\th}{\bm{\theta}}

\renewcommand{\x}{\mathbf{x}}

\renewcommand{\l}{\mathbf{l}}

\renewcommand{\q}{\mathbf{q}}
\renewcommand{\s}{\mathbf{s}}
\renewcommand{\t}{\mathbf{t}}

\renewcommand{\A}{\mathbf{A}}
\renewcommand{\B}{\mathbf{B}}

\renewcommand{\Q}{\mathbf{Q}}

\usepackage[table]{xcolor}
\usepackage{capt-of}

\newcommand{\rev}[1]{#1}

\begin{document}

\twocolumn[{
\maketitle
\vspace{-2em}
  \begin{center}
    \includegraphics[width=\textwidth]{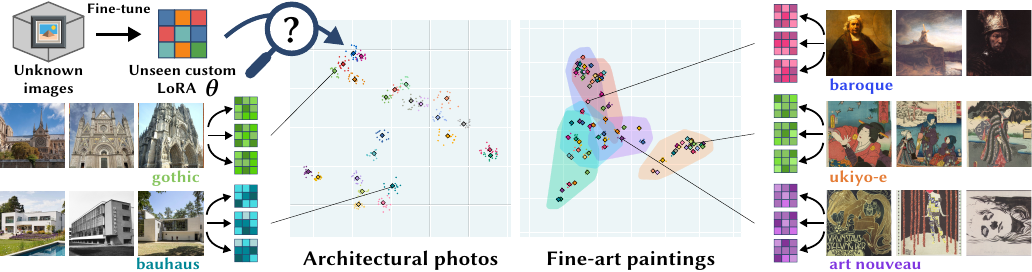}
    \captionof{figure}{
    We show that LoRA weights trained on stylized datasets serve as standalone feature representations: embeddings for architectural photos \cite{xu2014architectural} (left) and historic artworks \cite{liao22,saleh2016wikiart} (right) enable accurate clustering and retrieval of \textit{unseen} LoRAs.
        Art genres are shown as shaded convex hulls:
        \textcolor[HTML]{8036E0}{art nouveau},
        \textcolor[HTML]{3047EF}{baroque},
        \textcolor[HTML]{03C6DB}{expressionism},
        \textcolor[HTML]{00AA79}{post impressionism},
        \textcolor[HTML]{DB3019}{renaissance}, 
         \textcolor[HTML]{E57C35}{ukiyo-e}.
        LoRAs are color-coded by style, with centroids ($\diamondsuit$).
        Only a subset of samples is shown for clarity.
    }
    \label{fig:teaser}
  \end{center}
}]

{\renewcommand{\thefootnote}{*}\footnotetext{Now at the University of Victoria.}}

\begin{abstract}
Low-Rank Adaptation (LoRA) has become a standard approach for parameter-efficient fine-tuning of image diffusion models, with over 200K models trained to replicate artistic styles and shared in online communities.
One way to analyze styles of LoRA models is to generate image samples and apply image-based features like CLIP.
However, this generation requires additional computation and prompt selection.
In this work, we propose analyzing styles directly in the LoRA weight space, bypassing these limitations while achieving better performance even on unseen LoRAs.
Since in-the-wild LoRAs lack ground truth model details, we design an experimental setup and show that LoRA weights alone serve as effective feature representations of adapted styles, robust to variations in fine-tuning seeds and image selection.
We extensively evaluate using SD1.5 (the most popular base model within online communities) across data domains (fine-art paintings, architectural photos) and diverse prompt settings, with a small-scale SDXL experiment further verifying generalization.
These evaluations confirm that our approach, under the selected aspects of variability, preserves fine-grained styles and outperforms image-based methods in clustering and retrieval of unseen LoRAs.

\end{abstract}

\section{Introduction}
\label{sec:intro}

\begin{wrapfigure}{r}{0.055\linewidth}
  \includegraphics[width=\linewidth,trim=0.25cm 0.14cm 0.0cm 0.3cm]{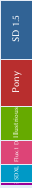}
\end{wrapfigure}
Among parameter-efficient fine-tuning (PEFT) methods for image generation models, Low Rank Adaptation (LoRA) \cite{hu21} has become a standard approach that allows low-cost customization of 
models to make them follow a specific style or subject from as few as 10–20 images.  
The adoption of LoRA in creative communities has been remarkable, with platforms like \href{https://civitai.com/}{Civitai}\footnote{A site for sharing, inferring, merging, and training text-to-image models. Statistics from May 2025.} hosting over 225K PEFT models, 97\% of those LoRAs. Stable Diffusion (SD) 1.5~\cite{rombach2022high,sd15} is the most popular base model, accounting for 31\% of all LoRAs, with ranks ranging from 2 to 384 and 16 and 32 being the most common.
These LoRA model weights found \textit{in the wild} are a new data format that represents artistic style.
Currently, there are no good tools to analyze, compare, and search among LoRAs.
The existence of such tools would be useful for exploration, attribution and crediting of style~\cite{ducru2024ai}, benefiting both artists who create original artwork and practitioners who train and share models.

\begin{figure}[t]
  \includegraphics[width=\linewidth]{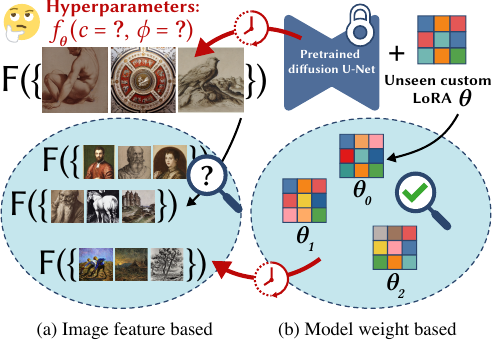}
  \caption{
    (a) Retrieving LoRAs based on image features requires additional image generation (red arrows), which rely on  hyperparameter choices.
    (b) Direct retrieval of LoRAs using the model weights $\th$ \textit{themselves} is accurate and avoids extra costs as well as the need for hyperparameters, such as prompts $c$.
  }
  \label{fig:vs_clip}
\end{figure}

One approach to analyzing the style of an existing LoRA model is to sample its output: generate images and analyze their style.
A stylized image $\x$ is generated by $f_{\th}(c, \phi)$ with model weights $\th$, a prompt $c$, and additional inference parameters $\phi$.
Analysis can then be performed on these images $\x$ by using pre-trained vision models (CLIP~\cite{radford21},  DINO~\cite{caron21}) or models fine-tuned for this purpose (AbC \cite{wang2023evaluating}, CSD \cite{somepalli2024measuring}).
These methods require not only extra computational cost but also non-trivial choices of prompt $c$ and parameters $\phi$ (Fig.~\ref{fig:vs_clip}a, red arrows), which can significantly alter the distribution of stylized images from that of real images (Fig.~\ref{fig:ad_quality}).

In this work, we show that LoRA weights $\th$ \textit{themselves} can potentially serve as standalone feature representations for the adapted styles, eliminating inference costs (Fig.~\ref{fig:vs_clip}b) and not requiring access to fine-tuning datasets, which are largely unavailable for in-the-wild LoRAs.
A controlled investigation of community-shared, in-the-wild LoRAs is challenging for a number of reasons: lack of ground truth, mismatching hyperparameters (subnetwork selection, rank, learning rate, training steps, etc.).
As a full investigation via controlled variation of all these factors would make early research in this emerging area difficult to initiate, we design an experimental setup that reflects key aspects of working with in-the-wild LoRAs.
In our experiments, all LoRAs are fine-tuned using the same base model (SD 1.5), training configuration, and rank (16).
To resemble the in-the-wild setup, we vary the selection and size of fine-tuning datasets and the random seeds --- we are interested in analyzing the pre-trained LoRAs without access to their underlying fine-tuning datasets at test time.
We show that the space of LoRA weights $\th$s clearly distinguishes fine-grained styles invariant to our selected aspects.
We propose methods to reduce the LoRA's dimensionality (Sec.~\ref{sec:subnetwork}) and improve retrieval accuracy for unseen LoRA queries (Sec.~\ref{sec:calibration}).

To quantitatively examine our approach, we construct two evaluation model databases
on distinct domains: fine-art \textit{paintings} (140 artists~\cite{liao22,saleh2016wikiart}) 
and architectural \textit{photos} (25 styles \cite{xu2014architectural}).
We 
evaluate our approach on style-based clustering and retrieval.
We compare to image-based features computed on the \textit{original} fine-tuning images (which are not shared in the wild): our method is on par or better.
When image-based features are restricted to \textit{generated} images, we greatly outperform.
We further experiment with SDXL~\cite{sdxl} (the second most popular base model, with variants comprising 25\% of Civitai LoRAs) and observe the same clustering behavior as SD1.5, along with small-scale experiments in two other domains (see appendix).

\section{Related Work}
\label{sec:related}

\rev{%
}

We begin by reviewing the prior arts of text-to-image generation, customization, and their connection to style. 
We then cover computational representations of style, and then  discuss weight-space representations more generally to motivate the use of LoRA weights as a data format.

\subsection{Text-to-image generation}

Recent advances in text-to-image generation have revolutionized content creation workflows.
Diffusion models~\cite{ho2020denoising, ho2022classifier, rombach2022high} have been particularly effective, 
and
the release of Stable Diffusion, a latent diffusion model~\cite{rombach2022high} trained on large-scale datasets, created a creative community around customizing these open source models via fine-tuning on specific artistic styles.
Dreambooth~\cite{ruiz2023dreambooth} and MultiDiffusion~\cite{kumari2023multi} present a set of techniques to customize diffusion models through fine-tuning, while the open-source community adopted Low Rank Adaptation (LoRA)~\cite{hu21} 
from the language model literature.
Several works augment the LoRA fine-tuning framework for different artistic applications. 
MuseumMaker~\cite{lin2024diffusion} adopts a continual learning to mix 
artistic styles in a streaming manner. 
ZipLoRA~\cite{shah2023ZipLoRA} proposes a method of mixing LoRA weights 
that takes into account subject-style interference. 
B-LoRA~\cite{frenkel2024implicit} 
finds LoRA weight components that correspond to style and subject for separate control. 
These works hint towards a connection between the weights of the LoRA and artistic style. 
In our work, we leverage this connection for the retrieval of style with unseen LoRAs.

\begin{figure}[t]
  \includegraphics[width=\linewidth]{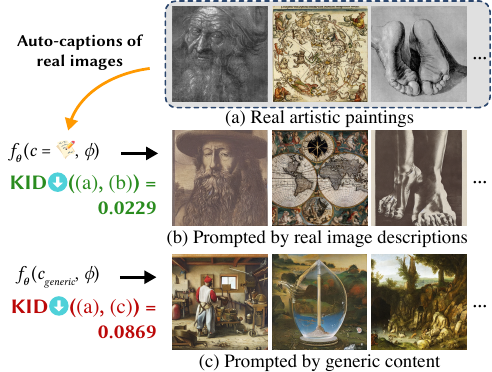}
  \caption{
    Prompt choices affect the distribution of generated images \cite{somepalli2024measuring}.
    For instance, descriptions of an artist's typical subjects produce images (b) with styles more closely matching real ones (a) than those from generic prompts (c), indicated by a lower Kernel Inception Distance (KID).
    However, information about real training images is rarely shared with the model, making (b) impractical.
  }
  \label{fig:ad_quality}
\end{figure}

\subsection{Style Representation}

The notion of style can be interpreted in several different ways, but we use the definition in art history: [style is] \textit{a distinctive manner which permits the grouping of works into related categories}, posed by Eric Fernie~\cite{fernie1995art, liu2023algorithms}. 
Humans excel at recognizing styles --- by recognizing brushstroke patterns, color palettes, and composition, humans can quickly distinguish between a painting by Van Gogh and Hokusai. 
To computationally emulate this capability, we frame style analysis as a clustering problem: we seek 
representations 
where artworks naturally group together in ways that mirror how humans categorize art by style.

Early methods extract low-level image statistics such as Fractal analysis~\cite{taylor1999fractal}, brush stroke wavelets~\cite{johnson2008image}, sparse coding~\cite{hughes2010quantification}, and color statistics~\cite{kim2014large} to quantify artistic styles.
Deep learning led to methods~\cite{karayev2013recognizing} that use pre-trained models to embed images, with applications like style-transfer~\cite{gatys2015neural, johnson2016perceptual}, perceptual image differencing~\cite{johnson2016perceptual, zhang2018unreasonable}, and style analysis~\cite{ruta2021aladin, ruta2022stylebabel, kumar2024introstyle}.
Later works use transformers~\cite{dosovitskiy2020image, caron21} to achieve improved performance.
Notably, CLIP~\cite{radford21} embeds images and corresponding textual labels into a shared embedding space, giving the space more semantic structure.
CLIP is a particularly effective embedding space for artistic works, with several works~\cite{balan2023ekila, wang2023evaluating, somepalli2024measuring} measuring the efficacy of CLIP and fine-tuned variants for artistic style analysis on real and generated images.
In contrast to works that extract style features from images --- where inference choices such as prompts can significantly affect the results (Fig.~\ref{fig:ad_quality}) --- our work \textit{directly} inspects the weight vectors of LoRAs fine-tuned on style images.
This is effective for 
analyzing collections of fine-tuned LoRAs (such as the ones shared online), as no additional inference (leading to stochasticity) 
is required.

\subsection{Weight Space Representations}
\label{sec:weight_space}

Using neural network weights directly
as an \textit{embedding} has been explored in various settings.
HyperNetworks~\cite{ha2016hypernetworks, alaluf2022hyperstyle, erkocc2023hyperdiffusion} use a neural network to generate the weights of another network.
Weights2weights~\cite{dravid24} 
explores the use of an embedding space of LoRAs (based on a large dataset of portrait photos) to generate LoRA weights and enable editing.
In contrast to these works that focus on creating a generative model in weight space for sampling and editing, we focus on inspecting LoRAs unseen to the database construction and evaluate on multiple, diverse data domains.

Several works also explore neural network weights as a \textit{data format}.
Luo et al.~\cite{luo2024stylus} recognize the proliferation of LoRA weights found in the wild to design a routing method to ensemble LoRAs from a database of LoRAs.
Lim et al.~\cite{lim2024learning} similarly recognize LoRAs as a new data format and 
uses LoRA weight vectors as input to predict performance metrics and data attribution.
We also treat LoRAs as a proliferating data format in its own right.
Other works explore emergent clustering behaviour within the weight-space of fine-tuned models.
Gueta et al.~\cite{gueta2023knowledge} measure distances between 95 full-weight fine-tuned language models and observe loose clustering within each of 3 coarse-grained NLP tasks.
We similarly show that there is clustering behavior within diffusion models LoRAs for \textit{styles} (analogous to \textit{tasks}) and further show that this can be leveraged for accurate clustering 
across a large database containing 3{,}360 models across 140 fine-grained styles.

A core assumption in treating the fine-tuned weights 
as an embedding 
is that the space of fine-tuned models is smoothly structured and semantically meaningful. 
Several works explore properties in which this assumption can hold for neural networks. 
Frankle et al~\cite{frankle20} show that when two models share the first $k$ training iterations (where $k$ is relatively small), they become \textit{linearly mode connected}, meaning that linear interpolations between fine-tuned models %
share similar performance.
This has 
inspired weight-space ensemble methods, to reduce overfitting \cite{wortsman22} and modify task-specific capabilities \cite{ilharco22}.
These concepts extend to LoRA weights, which all start from a pre-trained model, satisfying the condition for linear mode connectivity.

\rev{
}

\section{Method}

\begin{figure}[t]
  \includegraphics[width=\linewidth]{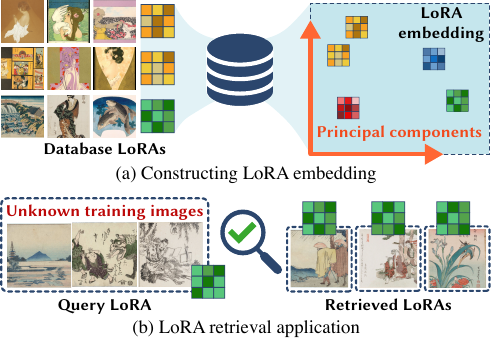}
  \caption{a
    (a) We construct an embedding of style-customized LoRAs via PCA.
    (b) This embedding enables applications such as retrieval using a query LoRA fine-tuned on unknown images, avoiding the need for additional image generation.
  }
  \label{fig:method}
\end{figure}

To analyze the style of unseen LoRAs, we introduce a method to construct style embeddings \textit{directly} from LoRA weights instead of the generated images (Fig.~\ref{fig:method}a).
We use PCA to project LoRA to a lower-dimensional space (Sec.~\ref{sec:pca}), explore the 
sub-network weights for computational efficiency   (Sec.~\ref{sec:subnetwork}), and introduce strategies to accurately retrieve unseen LoRAs (Fig.~\ref{fig:method}b, Sec.~\ref{sec:calibration}).

\subsection{Low-Rank Adaptation (LoRA)}

Our method constructs style embeddings from LoRA weights. LoRA is 
used to fine-tune pre-trained models to a downstream task based on a task-specific training set. 
The fine-tuning involves updating the 
linear weight matrices $ \mathbf{W}_0 \in \mathbb{R}^{m \times n} $ with update matrices $ \Delta \mathbf{W} $ such that the forward pass is $h = (\mathbf{W}_0 + \Delta \mathbf{W}) x$.
The update is applied to all %
matrices across the network. 
Observing that $ \Delta \mathbf{W} $ typically has a low intrinsic rank, LoRA reparameterizes this update matrix as $ \Delta \mathbf{W} = \mathbf{B} \mathbf{A} $, where $ \mathbf{B} \in \mathbb{R}^{m \times r} $, $ \mathbf{A} \in \mathbb{R}^{r \times n} $, and $ r \ll \min(m, n) $ is a small rank chosen by the user. The forward pass becomes:
\begin{equation}
h = \mathbf{W}_0 x + \frac{\alpha}{r} \, \mathbf{B}\mathbf{A} x,
\end{equation}
where $\alpha$ is a hyperparameter.
By updating only $ \mathbf{B} $ and $ \mathbf{A} $ during fine-tuning, LoRA significantly reduces the number of parameters, resulting in 
lower training
and storage costs. 

We then flatten the new matrix components of the LoRA and concatenate them to obtain the LoRA weight vector $\th$. Given $\text{vec}(\mathbf{A}) = [a_{11}, ..., a_{rn}]^\top \in \mathbb{R}^{r n \times 1}, 
\text{vec}(\mathbf{B}) = [b_{11}, ..., b_{mr}]^\top \in \mathbb{R}^{m r \times 1}$,
\begin{align*}
\th &= [\text{vec}(\mathbf{A}); \text{vec}(\mathbf{B})] \in \mathbb{R}^{(r n + m r) \times 1}
\end{align*}
The final $\th$ concatenates all $\B, \A$ across the entire network.
We use the LoRA weights fine-tuned on image sets of various styles to construct a database of LoRA weight vectors.

\begin{figure*}[t]
  \includegraphics[width=\linewidth]{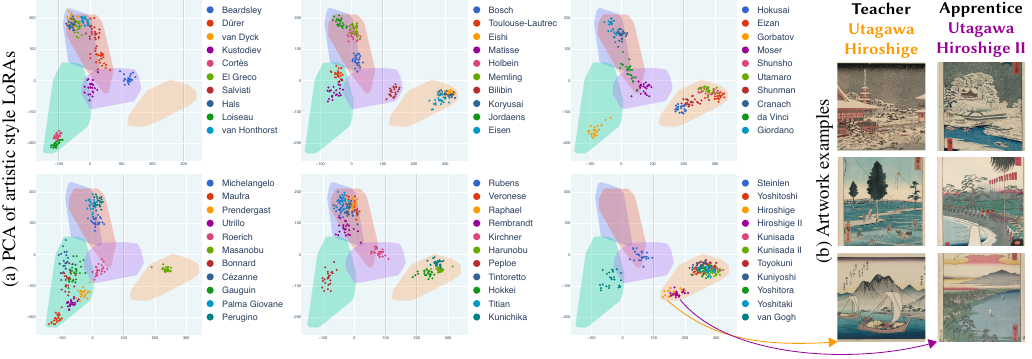}
  \caption{
    (a) Plot of our artistic style LoRA embedding along the first two PCs, with the same genre convex hulls as Fig.~\ref{fig:teaser}.  
    We display 63 ArtBench artists and split subsets across subfigures to reduce clutter.
    (b) The embedding aligns with art historical knowledge; for instance, a teacher-apprentice pair with similar styles appears close.
  }
  \label{fig:embedding}
  \vskip -0.3cm 
\end{figure*}

\subsection{Constructing Style Embeddings}
\label{sec:pca}

To inspect our style embedding approach, we construct LoRA databases from a collection of artists (or architectural styles) and their body of work by sampling images from each artist, fine-tuning a LoRA weight, and applying PCA to obtain lower-dimensional embeddings for analysis.  
We refer to LoRAs used to construct our database embeddings as \textit{training LoRAs} and LoRAs that are projected into this space (to query from the database) as \textit{test LoRAs}.
We refer to training individual LoRAs on image sets as \textit{fine-tuning}.
In this section, we focus on artistic paintings for clarity and show generalization to architectural photos in Sec.~\ref{sec:experiments}.

We build our style embedding based on ArtBench \cite{liao22} and WikiArt \cite{saleh2016wikiart}, datasets of public domain 
historical artworks labeled with authorship and genre (examples in Fig.~\ref{fig:teaser}, \ref{fig:method}).
We focus our experiments on SD1.5 due to its widespread use and the abundance of existing models.
For an artist $i$ among the $n=140$ 
artists with a sufficient number of artworks, we randomly sample 15 images from the training set to fine-tune a LoRA, yielding an unraveled weight vector $\th_{i,j} \in \mathbb{R}^{d}$, where $d = $~5{,}984{,}256 in our setting (Fig.~\ref{fig:method}a; see Sec.~\ref{sec:implementation} for experiment details).

To isolate the effect of LoRA weights from the text encoder, we fine-tune only the U-Net without token optimization.
Each LoRA is assigned a unique random identifier token \texttt{[V]}, prepended to an auto-caption for each image.
We manually verify that these auto-captions are almost always free of artist-related information, with rare name variants for 8 of 63 ArtBench artists appearing in only 1.6\% of prompts on average.
We fine-tune to collect a dataset $\mathcal{D} = \{\th_{i,j} \}$ with 30 LoRAs per artist.

To obtain lower-dimensional embeddings, we apply PCA to $\mathcal{D}$ and obtain the principal components (PCs) $\{\mathbf{q}_1, ..., \mathbf{q}_k\}$, resulting in Fig.~\ref{fig:embedding}.
We find that the weight vectors of LoRAs trained on images by the \textit{same} artist cluster together closely, despite their training image sets overlapping by only 37\% on average and 53\% at most, while LoRAs trained on images by \textit{different} artists tend to be well-separated (quantitative evaluations in Sec.~\ref{sec:clustering}).
We observe that the subspace of the first two PCs aligns with art history prior knowledge (Fig.~\ref{fig:embedding}, discussions in the appendix).

\subsection{Sub-Network Weights}
\label{sec:subnetwork}

While the 
LoRA method 
already 
reduce the number of parameters from the full network weights,
the 6M parameters
results in high computational costs.
Applying LoRA to a subset of the network, often cross-attention layers, is a common practice for efficient fine-tuning, adopted by both practitioners \cite{cloneofsimo} and in research \cite{kumari2023multi,shah2023ZipLoRA,dravid24,frenkel2024implicit}.
We ablate the style representation capabilities of different components under our training configuration.
We compare the clustering performance on the validation set of both our databases (we defer to Sec.~\ref{sec:clustering} for the experimental setup). 
Let $d$ be the dimension of the full-network LoRA vectors.
We report clustering metric ARIs ($\uparrow$) and corresponding dimensionalities on our 140-artist database: full network \{90.2\%, $d$\}, feedforward neural networks \{96.5\%, $0.47d$\}, self-attention \{94.8\%, $0.27d$\}, and cross-attention \{67.3\%, $0.27d$\}, with a similar ordering on architecture dataset.
\rev{These results suggest that cross-attention encodes less style than self-attention or feedforward layers, as prior work shows it mainly reflects prompt semantics while self-attention captures spatial structure~\cite{tumanyan2023plug,liu2024towards}, and because our LoRAs are trained with random \texttt{[V]} tokens instead of style words, cross-attention weights are unlikely to capture consistent style.}
Notably, the feedforward and self-attention layers demonstrate better representation capability than the full network.
In our evaluations, we use feedforward neural network weights for our weight vectors.

\subsection{Accurate Retrieval Given Unseen LoRAs}
\label{sec:calibration}

\begin{figure*}[t]
  \includegraphics[width=\linewidth]{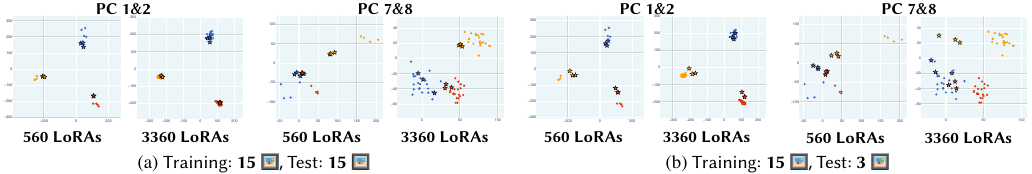}
  \caption{
  PCA plot of the training LoRAs ($\mathbf{\bullet}$) and test LoRAs ($\bigstar$). 
  With a smaller set of training LoRAs (560 LoRAs), the embedding for the training and test LoRAs drift apart. 
  While using a larger set of training LoRAs (3360) improves the drift, 
  the drift 
  is still apparent for test LoRAs fine-tuned under different settings (15 fine-tuning images for training LoRAs and 3 for test LoRAs).}
  \label{fig:drifts}
\end{figure*}

Our goal is to 
retrieve unseen test LoRAs from our training LoRA database.
Using the PCA naively is suboptimal,
as the PCA constructed from the training LoRAs is not robust enough to 
generalize to the test LoRAs.
The projection errors are caused by the limited amount of data fed to the PCA computation with respect to the high dimensionality of data ($d = $~5{,}984{,}256 or 2{,}795{,}520 with feedforward network), especially for the later PCs, which are more challenging to estimate accurately (see Fig.~\ref{fig:drifts} for the drifts observed).
We find that in the embedding space, test LoRAs (stars in Fig.~\ref{fig:drifts}) stay clustered together (within artist labels),
despite drifting away from corresponding training LoRA clusters.
This implies a simple transformation can correct the drifts from the training LoRA clusters.
We model the projection of test LoRAs into the training embedding as a variation of domain adaptation~\cite{pan2008transfer,fernando2013unsupervised,hendy2024tl} and correct the errors using an affine transformation with scaling $\s = \{s_{1, ..., k} \}$ and translation $\t = \{t_{1, ..., k} \}$ (exact calculations in the appendix).

We explore two different scenarios: training and test LoRAs may have overlapping or non-overlapping fine-tuning image sets with varying image set sizes.
Our experiments (Fig.~\ref{fig:drifts})
show that the translation $t_k \approx 0$ when projecting test LoRAs fine-tuned on datasets of the same size as the training LoRAs, regardless of whether their %
image sets overlap (Fig.~\ref{fig:drifts}a); and $t_k \neq 0$ when the sizes vary (Fig.~\ref{fig:drifts}b).
We propose two strategies to 
for retrieval
without the costly generation of large
LoRA databases:
\begin{enumerate}
\item{When the fine-tuning image set sizes are the same ($t_k \approx 0$), we use standard \textit{cosine} distance.
This distance is also found effective for full-weight fine-tuned LLMs~\cite{gueta2023knowledge}.}
\item{Only when the image set sizes differ, we apply an additional calibration step as it requires additional LoRA construction. Instead of computing on test LoRAs, we construct calibration LoRAs with known labels ($m_{\text{cali}}$ per artist) to solve for the affine transformation $\s, \t$.}
\end{enumerate}

\section{Experiments}
\label{sec:experiments}

\begin{figure*}[t]
  \includegraphics[width=\linewidth]{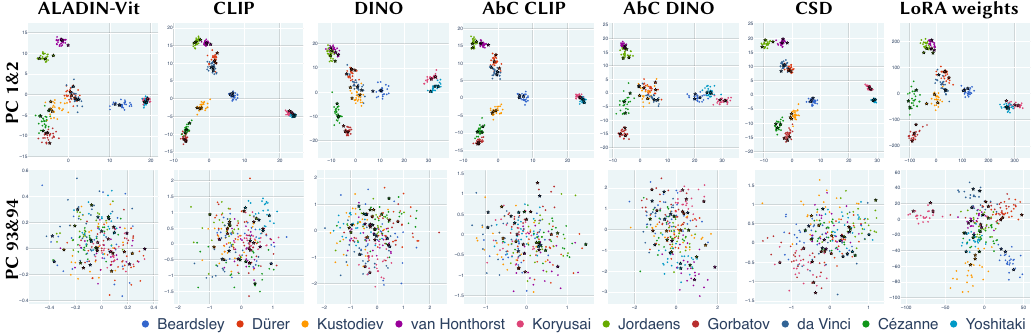}
  \caption{
    PCA visualizations of image features and LoRA weights (ours).
    The embeddings from different approaches are similar along the first few PCs, but only the LoRA weights maintain distinct style clusters in the later axes while others become more entangled.
    For clarity, we select only 10 artists with training samples ($\mathbf{\bullet}$) and test samples ($\bigstar$) visualized.
  }
  \label{fig:clustering}
    \vskip -0.6cm 
\end{figure*}

We conduct a series of experiments to evaluate how well our approach can analyze unseen LoRAs (Sec.~\ref{sec:implementation}).
We first show that unsupervised embedding of LoRAs creates meaningful style clusters, outperforming baselines (Sec.~\ref{sec:clustering}), and then evaluate retrieval from this embedding (Sec.~\ref{sec:retrieval}).

\begin{table*}[t]
  \scriptsize
\caption{
    Quantitative clustering metrics.
    The training samples consist of data used to construct the embeddings.
    The test samples are data not seen during the embedding construction.
    Higher is better (mean ± std over 10 runs); \fcolorbox{ForestGreen!25}{ForestGreen!25}{best} and \fcolorbox{ForestGreen!10}{ForestGreen!10}{second best} values are highlighted.
}
\centering
\label{tab:clustering}
\begin{tabular*}{\textwidth}{@{\extracolsep{\fill}}l|cccc|cccc}
\hlineB{3}
\multirow{3}{*}{Methods} & \multicolumn{4}{c|}{Artistic Dataset} & \multicolumn{4}{c}{Architectural Dataset} \\
  & \multicolumn{2}{c}{ARI $\uparrow$ (\%)} & \multicolumn{2}{c|}{NMI $\uparrow$ (\%)} & \multicolumn{2}{c}{ARI $\uparrow$ (\%)} & \multicolumn{2}{c}{NMI $\uparrow$ (\%)} \\
 & Train & Test & Train & Test & Train & Test & Train & Test \\
\hlineB{3}
ALADIN-ViT & $88.0_{\pm 2.2}$ & $85.1_{\pm 2.4}$ & $97.6_{\pm 0.4}$ & $98.2_{\pm 0.3}$ & $79.8_{\pm 3.7}$ & $76.4_{\pm 2.0}$ & $91.6_{\pm 1.4}$ & $94.6_{\pm 0.5}$ \\

CLIP & $89.9_{\pm 1.5}$ & $86.7_{\pm 1.9}$ & $98.2_{\pm 0.3}$ & $98.5_{\pm 0.2}$ & \cellcolor{ForestGreen!10} $93.9_{\pm 4.3}$ & \cellcolor{ForestGreen!10} $89.1_{\pm 5.1}$ & \cellcolor{ForestGreen!10} $98.5_{\pm 1.0}$ & \cellcolor{ForestGreen!10} $97.8_{\pm 0.8}$ \\

DINO & $87.9_{\pm 1.1}$ & $84.7_{\pm 1.5}$ & $97.6_{\pm 0.2}$ & $98.1_{\pm 0.3}$ & $91.6_{\pm 3.3}$ & $87.7_{\pm 4.3}$ & $97.8_{\pm 0.7}$ & $97.7_{\pm 0.7}$ \\

AbC CLIP & $89.5_{\pm 1.8}$ & $86.9_{\pm 2.4}$ & $98.2_{\pm 0.3}$ & $98.6_{\pm 0.3}$ & $90.9_{\pm 4.0}$ & $86.8_{\pm 4.4}$ & $97.6_{\pm 1.1}$ & $97.4_{\pm 0.8}$ \\

AbC DINO & \cellcolor{ForestGreen!10} $91.5_{\pm 1.7}$ & \cellcolor{ForestGreen!10} $89.0_{\pm 2.2}$ & $98.5_{\pm 0.3}$ & \cellcolor{ForestGreen!10} $98.8_{\pm 0.2}$ & $90.3_{\pm 4.5}$ & $85.4_{\pm 6.1}$ & $97.1_{\pm 1.4}$ & $96.9_{\pm 1.4}$ \\

CSD & $91.5_{\pm 1.8}$ & $88.6_{\pm 2.5}$ & \cellcolor{ForestGreen!10} $98.6_{\pm 0.2}$ & $98.8_{\pm 0.2}$ & $91.8_{\pm 4.3}$ & $86.8_{\pm 5.6}$ & $98.0_{\pm 0.9}$ & $97.5_{\pm 0.8}$ \\
\hlineB{1}

Ours & \cellcolor{ForestGreen!25} $97.3_{\pm 1.5}$ & \cellcolor{ForestGreen!25} $96.5_{\pm 2.0}$ & \cellcolor{ForestGreen!25} $99.6_{\pm 0.2}$ & \cellcolor{ForestGreen!25} $99.7_{\pm 0.1}$ & \cellcolor{ForestGreen!25} $99.5_{\pm 1.6}$ & \cellcolor{ForestGreen!25} $99.3_{\pm 2.2}$ & \cellcolor{ForestGreen!25} $99.9_{\pm 0.4}$ & \cellcolor{ForestGreen!25} $99.9_{\pm 0.4}$ \\

\hlineB{3}
\end{tabular*}
\end{table*}

\begin{figure*}[t]
  \includegraphics[width=\linewidth]{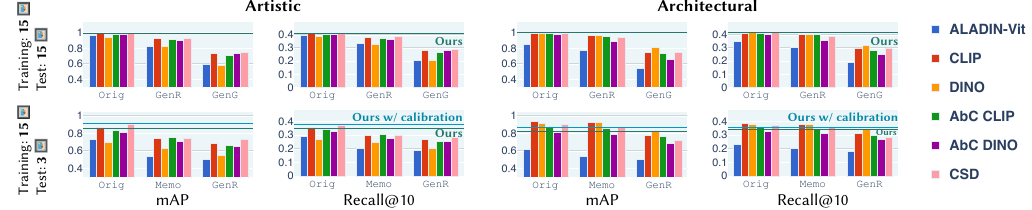}
  \caption{
    Retrieval metrics. 
    Our results are shown with a horizontal bar, as they produce the same metrics across all scenarios.
    The baseline methods are affected by the images used: original fine-tuning images (\texttt{Orig}), images prompted with real image descriptions (\texttt{GenR}), images prompted with generic prompts (\texttt{GenG}), images memorized by LoRAs (\texttt{Memo}).
  }
  \label{fig:retrieval_stats}
  \vskip -0.6cm 
\end{figure*}

\subsection{Experiment Settings}
\label{sec:implementation}

We refer to LoRAs used to construct our database embeddings as \textit{training LoRAs} and those projected into this space for querying as \textit{test LoRAs}.
To distinguish from the LoRA embedding construction, we refer to training individual LoRAs on image sets as \textit{fine-tuning}.

\textbf{Databases~}
We construct two evaluation LoRA databases: Artistic (fine-art paintings from ArtBench \cite{liao22}, WikiArt \cite{saleh2016wikiart}; each artist is treated as having a single style) and Architectural (architectural photos \cite{xu2014architectural}).
We select sufficiently large original images ($\text{min}(w,h)\!\geq\!512$), resize, and center-crop to $512\!\times\!512$ for SD1.5 LoRA fine-tuning.
For the Artistic database, we pick artists from ArtBench then from WikiArt without duplication.
Styles are selected if they have a sufficient number of samples ($\geq\!100$), resulting in $n=140$ artists across 14 genres, and $n=25$ architectural styles.
We auto-caption preprocessed images with the LAVIS \cite{li2022lavis} implementation of BLIP \cite{li2022blip}.

We inspect our databases with clustering as a test case, using test LoRAs fine-tuned on image sets that \textit{partially overlap} with but distinct from training LoRAs.
For our retrieval test, we reuse the same training LoRAs.
We hold out 30 images per style, ensuring query LoRAs are fine-tuned on images \textit{fully disjoint} from those of database LoRAs.

\textbf{Baselines~}
We compare against common image features: style descriptors (ALADIN-Vit \cite{ruta2022stylebabel}), self-supervised (DINO \cite{caron21}), vision-language (CLIP \cite{radford21}) and fine-tuned image attribution (AbC CLIP, AbC DINO
~\cite{wang2023evaluating}, CSD \cite{somepalli2024measuring}).
Since LoRAs are trained on a \textit{set} of images, 
for fair comparison, we compute a 
single embedding
for the set 
by taking the average over encoded image features as in previous work~\cite{ruta2022stylebabel,somepalli2024measuring}.
Images $\x_j$ can be original or generated, depending on the experiment.
We verify that our LoRAs are properly fine-tuned for specific styles in the appendix.

\subsection{Clustering}
\label{sec:clustering}

We first evaluate how well samples cluster according to style in the baseline and weight-space embedding spaces.
We apply k-means where $k$ as the number of styles in the database (140 and 25 for artistic and architectural, respectively)
to embeddings of training samples.
We split the 30 LoRAs per style into 24 training, 3 validation and 3 test samples for both data domains.
For this clustering experiment, the baselines are given real images from the LoRA fine-tuning sets.
We measure common clustering metrics: Adjusted Rand Index (ARI) and Normalized Mutual Information (NMI).
Each reported number is averaged over 10 runs with different seeds.

\textbf{Visual comparison of PCA embeddings~}
We show PCA embedding visualizations of both the baselines and our method in Fig.~\ref{fig:clustering}.
Qualitatively, the embeddings 
(for both the baselines and our method)
of artwork or photo subsets from the same style should be closer to each other than to those from different styles.
We observe that embeddings from 
all
approaches exhibit similarities along the first 
two principal components (PCs).
However, as we examine the later PCs, 
our
LoRA embedding continues to produce well-separated style clusters up to the 94th 
PCs,
while the baselines
become more entangled.
Interestingly, LoRA weight style embeddings show a strong structural alignment, qualitatively and quantitatively, with image feature embeddings, suggesting a meaningful correspondence across different style embeddings (see appendix for metrics).

\textbf{Quantitative clustering performance~}
We report ARIs and NMIs in Table~\ref{tab:clustering}.
Our approach generalizes well, achieving similar performance on both the training and test sets and outperforming all baselines.
Our approach also maintains high accuracy across both artistic paintings and photos, while baselines trained specifically for \textit{artistic} styles (ALADIN, AbC, CSD) struggle on photo data.

\begin{wrapfigure}{r}{0.22\linewidth}
  \includegraphics[width=\linewidth,trim=0.8cm 0.5cm 0.0cm 0.52cm]{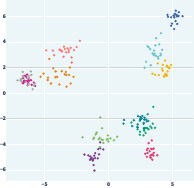}
\end{wrapfigure}
\textbf{SDXL Verification~}
We construct a small-scale dataset (12 artists, 20 LoRAs each) using SDXL as the base model.
We qualitatively verify that SDXL LoRAs cluster by adapted style in the PCA embedding space (inset).
Like image features, we observe similarity along the first two PCs between SD1.5 and SDXL LoRAs (details in the appendix).

\subsection{Retrieval with Unseen LoRAs}
\label{sec:retrieval}

We conduct a retrieval experiment that matches unseen LoRAs to their corresponding training LoRAs of the same style.
Given a query LoRA fine-tuned on images of a single artistic or architectural style, we retrieve the top matches using k-nearest neighbors, using cosine distance for all baselines.
We measure two retrieval metrics: mean Average Precision (mAP) and Recall@k (we report $k=10$, with 24 relevant LoRA instances per style in our training database).
Each number is averaged over 5 queries per style, totaling 700 artistic and 125 architectural queries.

\textbf{Baseline evaluation~}
Evaluating this LoRA retrieval task for image embedding based baselines requires additional \textit{query images} which are associated with the LoRA (in contrast to our weight-space method that can directly query the database with the LoRA).
We use three types of query images with decreasing prior knowledge about the LoRAs and their training sets, progressively aligning with real-world scenarios.
(1) \textit{Original}: The LoRA's training set is known, allowing features to be computed from the original images.
(2) \textit{Generated with real descriptions}: The querier has knowledge of the ground-truth (GT) style, enabling them to generate prompts with content typical of the GT.
(3) \textit{Generated with generic prompts}: The querier has no knowledge of the style and must use prompts of generic content; we use ChatGPT-generated prompts for Artistic and ``a building'' for Architectural.
The latter two settings require additional image generation (Fig.~\ref{fig:vs_clip}) and are validated in the appendix.
As far as we know, in current online communities, LoRAs are rarely shared with their training images, making the two \textit{generated} scenarios more realistic.

\textbf{Experimental settings~}
First, we test unseen LoRAs where the fine-tuning configs are the same as the training LoRAs (15 fine-tuning images), but the image sets are \textit{fully disjoint} from the fine-tuning image set (Fig.~\ref{fig:retrieval_stats}, row 1).
Additionally, we test query LoRAs fine-tuned on image sets that are \textit{fully disjoint} from the training LoRA image sets and \textit{also} have a different dataset size of 3 (Fig.~\ref{fig:retrieval_stats}, row 2).
This scenario arises when the querier lacks enough images to fine-tune a high-quality LoRA and seeks a better pre-trained alternative.
In this 3-image setting, to ensure fairness, we %
prompt the test LoRAs with the exact auto-captions used for fine-tuning (unknown in practice), resulting in generated images that are memorized versions of the originals.
The number of query images for baseline methods is adjusted accordingly.
These scenarios for the baseline methods affect query and database respectively.
We report results for settings where query and database prompting match, and see the appendix for additional combinations.

\textbf{Using original fine-tuning images~}
for both query and embedding serves as a stress test for our approach, as baseline methods are specifically designed for this setting.
Comparing image-based feature performance with 15 and 3 query images confirms that aggregating more queries improves results, while a query size of 3 is more challenging.
As shown under \texttt{Orig} in Fig.~\ref{fig:retrieval_stats}, our approach is on par with all baselines when retrieval is easier (query size 15) and performs slightly better in more challenging cases (bottom left, Fig.~\ref{fig:retrieval_stats}).
Although applying our approach to image feature extraction is impractical due to the  LoRA fine-tuning cost, this test shows that our method can outperform methods specifically designed for this task and remains robust even under challenging conditions, such as when database and query LoRAs share no fine-tuning images.

\textbf{Using generated images~}
worsens the baselines as expected, matching quality measures in the appendix.
In practice, beyond the decreased performance, generating and storing these images also incurs significant costs (roughly 210 GPU hours with NVIDIA Titan RTX, 27 GB for our Artistic database).
Our approach achieves significantly better performance in both mAP and Recall, avoiding the uncertainties and costs of inference, except in the 3-image setting with memorized images for architectural photos, where CLIP, DINO perform well (bottom right, Fig.~\ref{fig:retrieval_stats}).

\textbf{Retrieval strategies~}
Even \textit{without calibration}, using cosine distance for retrieval yields better performance than the image-based approaches with prior knowledge (\texttt{Memo} and \texttt{GenR}).
Calibration enhances accuracy by approximately 5\% (blue lines, Fig.~\ref{fig:retrieval_stats}), although it requires additional fine-tuning with a similar fine-tuning set configuration.
Here, both calibration and test LoRAs are fine-tuned on 3 images that are disjoint from each other and the training LoRAs.

\subsection{Ablations}
\label{sec:ablations}

\textbf{PCA~}
We compute 160 and 100 PCs for the Artistic and Architectural LoRA databases, using the same PCA for both tasks. 
Grid search on the clustering validation set yields optimal dimensions of 149 and 23, which roughly match the number of styles (140 and 25).

\begin{wrapfigure}{r}{0.26\linewidth}
\includegraphics[width=\linewidth,trim=0.6cm 0.5cm 0.0cm 0.4cm]{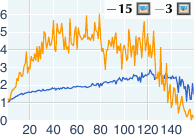}
\end{wrapfigure}
\textbf{Calibration}
The performance improvement on test sets for retrieval is presented in Fig.~\ref{fig:retrieval_stats} (``Ours'' vs. ``Ours w/ cali'').
Empirically on Artistic, we observe that $s_1 \approx 1$; $s_{k_1} < s_{k_2}$ when ${k_1} < {k_2}$ until approximately PC 120 and 80 for projecting 15-image-set and 3-image-set LoRAs respectively, when the projections become too inaccurate for effective calibration (inset).
To ablate our choice of using cosine distance when projecting unseen LoRAs fine-tuned on 15 images (the same as our database LoRAs), we report $t_k < 4.5\times 10^{-3} \approx 0$ under min-max normalization per PC.

\section{Limitations and Future Directions}
\label{sec:discuss}

\textbf{Limitations~}
Our eventual goal is to research style directly on LoRAs downloaded from model-sharing sites like Civitai.
In-the-wild LoRAs are often trained in diverse conditions with increasing difficulties to analyze: (1) varying ranks; (2) different training steps, learning rates, etc.; (3) fine-tuning image sets of unknown and mixed authorship; (4) different pre-trained checkpoints.
Across all our experiments, all LoRAs share the same dimensionality (e.g., rank or sub-network selection).
We conduct preliminary experiments with various-rank LoRAs and observe consistent results: grouping LoRAs by rank, constructing PCA embeddings within each group, and aligning them by sign yields embeddings with high structural similarity (see appendix).
While (4) breaks the linear mode connectivity assumptions, \citet{ainsworth2022git} show that this could be resolved after finding certain permutations.
To isolate our findings, we disabled text encoder fine-tuning, but in the future, its effect on LoRA styles should be included.
Our experiments are on a modest image diffusion model (SD1.5), still the most popular choice for practitioners on Civitai (probably due to its low cost), with a small-scale verification experiment on SDXL.
We expect our findings to generalize to larger models (SD3, Flux) but leave this for future work given the high training cost ($\sim$5h per Flux LoRA).

\textbf{Future applications~}
As larger models like Flux (currently accounting for only 13\%) become standard for practitioners, we see an increased need for coping with rank, training, and architectural differences.
One interesting direction is zero-shot LoRA fine-tuning: e.g., determining how to combine LoRAs from a database given some training images (generalizing our \texttt{Orig} setting).
A recent work \cite{jin2024retrieval} 
merges LLM LoRAs indirectly by computing blending weights in training data feature space.
Our findings show image-feature spaces and LoRA weight spaces are deeply related, but not by a simple similarity transform (as \citet{jin2024retrieval}~imply). 
Similarly, moving beyond image-based features, the task of attribution~\cite{zheng2023intriguing, georgiev2023journey, wang2024attributebyunlearning} could be conducted directly with LoRA weights,
following the emerging practice of training, sharing, and merging LoRAs.

\textbf{Broader impacts~}
We envision our work to be a step toward a mutually beneficial relationship between professional artists and generative technology.
While not extending generative \emph{capabilities}, our work offers an alternative lens to analyze genres and styles.
For example, this could support style quantification and comparison tools that help artists develop their personal styles.
Unauthorized style mimicry within online model sharing communities is a major concern of artists~\cite{honig2024adversarial}, not fully resolved
by data ``cloaking'' and poisoning ~\cite{shan2023glaze,shan2024nightshade}.
While our method may aid retrieval of unauthorized models, we hope improved style analysis and attribution tools will foster a more trustworthy, open, and mutually beneficial text-to-image ecosystem.

{
    \small
    \begingroup
    \let\l\relax

    \endgroup
}

\clearpage
\appendix

\section{Implementation Details}
\subsection{LoRA Fine-tuning}
We utilize Stable Diffusion 1.5 \cite{rombach2022high,sd15} as our pre-trained base model for our main results.
We conduct LoRA fine-tuning using a popular online repository \cite{sdscripts}.
The fine-tuning objective contains the denoising term and DreamBooth fine-tuning term \cite{ruiz2023dreambooth}.
We restrict fine-tuning to U-Net, including all linear weights in self-attentions, cross-attentions, and feedforward neural networks.
Our default setting uses rank $r = 16$, $\alpha=8$, and 1760 fine-tuning steps with learning rate $5\times 10^{-4}$ and Adam optimizer \cite{KingBa15}.

For the small scale SDXL experiment, we use SDXL 1.0 \cite{sdxl} as the base model.
Our SDXL setting uses rank $r = 1$, $\alpha=1$, and 1050 fine-tuning steps with learning rate $1\times 10^{-4}$ and Adam optimizer \cite{KingBa15}, with the other setup the same as the SD1.5 experiments.

\subsection{Compute Details}
We conduct our experiments on four NVIDIA Titan RTX 24GB cards and one NVIDIA RTX A6000 48GB.
The preliminary experiments take roughly 320 hours (training one SD1.5 LoRA on one NVIDIA Titan RTX takes around 10 minutes with our configuration).
For the final experiments, the main SD1.5 experiments take approximately 700 hours, and the SDXL experiment takes approximately 120 hours (training one SDXL LoRA on one NVIDIA Titan RTX takes around 30 minutes with our configuration).

\subsection{Additional Details}
We apply PCA using the IncrementalPCA implementation of scikit-learn~\cite{scikit-learn}.
For metric computation, we compute mAP using the formula $\text{mAP} = \sum_{i=1}^Q \text{AP}(q_i) / Q$, where $Q$ is the number of queries and $\text{AP}(q_i) = \sum_{k=1}^{N} \text{P@}k \cdot \text{rel}(k)$ is the average precision for query $q_i$; here, $\text{P@}k = \frac{1}{k} \sum_{j=1}^{k} \text{rel}(j)$ is the precision at rank $k$, and $\text{rel}(k) \in \{0,1\}$ indicates whether the item at rank $k$ is relevant. 
Recall@k is defined as the proportion of relevant items retrieved in the top $k$ results, computed as $\text{Recall@}k = \sum_{j=1}^{k} \text{rel}(j) / R$, where $R$ is the total number of relevant items for the query and $\text{rel}(j) \in \{0,1\}$ indicates whether the item at rank $j$ is relevant.
KID with torchmetrics implementation~\cite{torchmetrics}.
All statistics (mean and standard deviation) are computed using numpy~\cite{harris2020array}.

\section{Generation Quality}

\subsection{Example Prompts and Images}

\begin{figure}[t]
  \includegraphics[width=\linewidth]{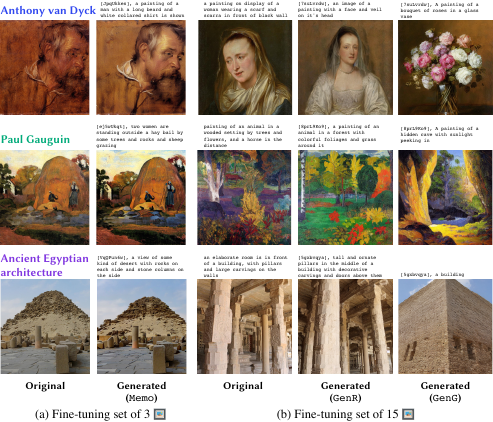}
  \caption{
  Example artworks and photos used for LoRA training (``Original''), memorized images generated using the auto-captions from fine-tuning (\texttt{Memo}), images generated using descriptions of the original images (\texttt{GenR}), and images generated using generic prompts (\texttt{GenG}).
  We show images generated by LoRAs fine-tuned on different image sets: 15 images (a) and 3 images (b).
  }
  \label{fig:example}
\end{figure}

We fine-tune each LoRA with a unique random identifier token \texttt{[V]}, images, and automatically generated captions.
Example training data is shown in Fig.~\ref{fig:example}.

We fine-tune LoRAs under two settings using training image sets of different sizes, 15 and 3 respectively.
For the comparison methods, we evaluate with three types of prompts, progressively aligning with real-world scenarios and with increasingly diverging from the original images: 
(1) \textit{Generated with real descriptions} (\texttt{Memo}): The generated image is a memorized version of the original fine-tuning image; we only use this for the 3-image-set case.
(2) \textit{Generated with real descriptions} (\texttt{GenR}): The querier knows the ground-truth (GT) artist and their real images, enabling them to generate prompts with content typical of the GT.
(3) \textit{Generated with generic prompts} (\texttt{GenG}): The querier has no knowledge of the style and must use prompts of generic content; we use ChatGPT-generated prompts for Artistic and ``a building'' for Architectural.

\subsection{Quantitative Validation}
\label{sec:quality}

\begin{wrapfigure}{r}{0.4\linewidth}
\includegraphics[width=\linewidth,trim=0.4cm 0.4cm 0.0cm 0.2cm]{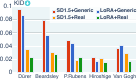}
\end{wrapfigure}
For our retrieval experiment, to ensure that query images generated by LoRAs capture the style of the fine-tuning images, we validate generation quality using KID$\downarrow$ metric~\cite{binkowski2018demystifying}, an unbiased metric for small real-image sample sizes.
We select artists with at least 200 real images, resulting in 11 artists, and randomly pick five training LoRAs per artist (five shown in inset).
We infer with two types of prompts: \textit{generic content} (red, blue) and \textit{real description} (yellow, green).
The former consists of 
500 simple prompts generated by ChatGPT; the latter reruns auto-captioning on the original images to obtain similar content while avoiding memorization.
We compare LoRA-generated images with \texttt{[V]} token (blue, green) with those from the base model prompted with ``in the style of \texttt{[artist name]}'' (red, yellow).
Unsurprisingly, \textit{real descriptions} yield significantly better KID measures on average (as small as 0.0237), highlighting the influence of prompt choices on image-feature-based comparisons.
Between the base model and LoRAs, LoRA-generated images achieve lower KIDs on average, indicating better alignment with the artist's style.
As extensively studied and reported~\cite{somepalli2024measuring}, the SD1.4 base model demonstrates strong performance in generating stylized images of well-known artists.
Although our base model (SD1.5) is different, we believe its quality is at least comparable, if not better, ensuring that LoRA fine-tuning and inference are properly conducted.

\section{Embedding}

\subsection{Embedding Visualization}

We observe that the subspace of the first two principal components aligns with art history prior knowledge (Fig.~\ref{fig:embedding}).
For instance, the eastern genre (\textcolor[HTML]{E57C35}{ukiyo-e}) is distinctly separated from the western genres.
Among the western genres, \textcolor[HTML]{8036E0}{art nouveau} is closest to \textcolor[HTML]{E57C35}{ukiyo-e}, reflecting the historical influence of \textcolor[HTML]{E57C35}{ukiyo-e} on \textcolor[HTML]{8036E0}{art nouveau} during its development, as well as the shared medium of printmaking between the two.
The overlapping artistic style clusters reflect known teacher-apprentice or influence relationships: \textcolor[HTML]{E57C35}{Utagawa Hiroshige, Utagawa Hiroshige II}, \textcolor[HTML]{636EFA}{Peter Paul Rubens, Anthony van Dyck}, and \textcolor[HTML]{DB3019}{Pietro Perugino, Raphael}.
Several close artistic style clusters (e.g., \textcolor[HTML]{DB3019}{da Vinci, Dürer, Michelangelo}) also match prior quantitative analyses of artworks \cite{bressan2008analysis}.

Similarly, we observe that the subspace of the first two principal components aligns with architectural style prior knowledge (Fig.~\ref{fig:teaser}, left).
Ancient Egyptian and Achaemenid being close reflects their roots in ancient monumental design principles, emphasizing symmetry, scale, and symbolic elements.
Baroque, Renaissance, and Gothic styles are relatively close to each other, which aligns with their historical progression and shared artistic principles. Gothic architecture heavily influenced Renaissance architecture, and Baroque emerged as a dramatic extension of Renaissance ideals.
American Craftsman and American Foursquare appear close, indicating their shared emphasis on regional identity in North America.

\begin{table*}[t]
  \footnotesize
\caption{
    The mean cosine similarity ($\uparrow, [-1, 1]$) between LoRA weight embedding and aggregated image feature embeddings derived from the same image set.
    `ALAD' refers to `ALADIN-Vit'.
}

\label{tab:sim}
\resizebox{\textwidth}{!}{
\begin{tabular*}{1.5\linewidth}{@{\extracolsep{\fill}}l|cccccc|l|cccccc@{\extracolsep{\fill}}}
\hlineB{3}
PC & ALAD & CLIP & DINO & AbC CLIP & AbC DINO & CSD &
PC   & ALAD & CLIP & DINO & AbC CLIP & AbC DINO & CSD \\
\hline
2 & 0.88 & 0.87 & 0.89 & 0.83 & 0.91 & 0.80 & 40 & 0.44 & 0.43 & 0.44 & 0.39 & 0.48 & 0.37 \\
3 & 0.87 & 0.81 & 0.83 & 0.79 & 0.90 & 0.73 & 50 & 0.42 & 0.40 & 0.42 & 0.36 & 0.45 & 0.35 \\
4 & 0.81 & 0.72 & 0.82 & 0.70 & 0.86 & 0.64 & 60 & 0.39 & 0.38 & 0.39 & 0.34 & 0.42 & 0.33 \\
5 & 0.76 & 0.73 & 0.77 & 0.68 & 0.81 & 0.64 & 70 & 0.37 & 0.36 & 0.37 & 0.33 & 0.39 & 0.32 \\
6 & 0.70 & 0.68 & 0.74 & 0.63 & 0.78 & 0.63 & 80 & 0.36 & 0.34 & 0.35 & 0.31 & 0.37 & 0.30 \\
7 & 0.67 & 0.66 & 0.71 & 0.60 & 0.76 & 0.58 & 90 & 0.34 & 0.32 & 0.33 & 0.30 & 0.35 & 0.29 \\
8 & 0.65 & 0.65 & 0.69 & 0.56 & 0.75 & 0.55 & 100 & 0.32 & 0.31 & 0.32 & 0.28 & 0.33 & 0.27 \\
9 & 0.65 & 0.62 & 0.66 & 0.54 & 0.72 & 0.54 & 110 & 0.31 & 0.29 & 0.30 & 0.27 & 0.31 & 0.26 \\
10 & 0.62 & 0.62 & 0.65 & 0.53 & 0.71 & 0.52 & 120 & 0.29 & 0.28 & 0.28 & 0.26 & 0.30 & 0.25 \\
20 & 0.53 & 0.51 & 0.53 & 0.46 & 0.58 & 0.44 & 130 & 0.28 & 0.27 & 0.27 & 0.25 & 0.29 & 0.24 \\
30 & 0.48 & 0.47 & 0.48 & 0.42 & 0.52 & 0.40 & 140 & 0.27 & 0.26 & 0.26 & 0.24 & 0.28 & 0.24 \\

\hlineB{3}
\end{tabular*}
}
\end{table*}

\subsection{Embedding similarity}
\label{sec:embed_sim}

\begin{wrapfigure}{r}{0.22\linewidth}
\includegraphics[width=\linewidth,trim=0.5cm 0.4cm 0.0cm 0.5cm]{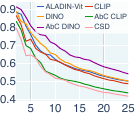}
\vskip -0.2cm
\end{wrapfigure}
We quantify the similarity between PCA embeddings by measuring the cosine similarity ($\uparrow, [-1, 1]$) between corresponding samples with increasing PCA dimensions (inset).
The results on Artistic (140 artists) show that embeddings with the first two PCs exhibit relatively high similarity ($[0.80, 0.91]$), but this similarity diminishes as additional PCs are included.
To ensure a meaningful comparison, we vary PC signs to optimally align the two embeddings.
The detailed statistics are in Table~\ref{tab:sim}.

\begin{figure}[t]
  \includegraphics[width=\linewidth]{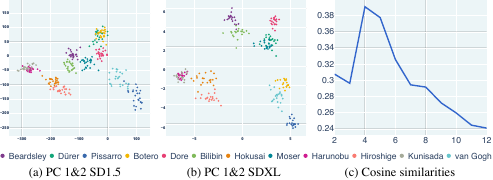}
  \caption{
  PCA embeddings of LoRA weights trained with SD1.5 (a) and SDXL (b) on the same 12 artists.
    SD1.5 embeddings are taken from the full set of 140 artists, while SDXL embeddings are computed on this subset only.
    (c) Cosine similarities with increasing PCA dimensions between the two embeddings show relatively weak alignment ($[0.24, 0.39]$).
    Some structure is preserved, though Botero (yellow) appears as an outlier.
  }
  \label{fig:sdxl}
\end{figure}

We also quantify the PCA embedding similarity between the SD1.5 and SDXL base models (Fig.~\ref{fig:sdxl}).
The SD1.5 embedding is the same as the one computed on 140 artists, while the SDXL embedding is computed on a smaller set of 12 artists.
We measure the similarity using the shared subset of 12 artists.
Compared to the similarity between SD1.5 and image-based features, the similarity between SD1.5 and SDXL embeddings is relatively weak ($[0.24, 0.39]$).
A closer investigation into the first two PCs reveals some structural similarity with Fernando Botero (yellow, Fig.~\ref{fig:sdxl}) as an outlier.

\section{Accurate Retrieval with Style Embeddings}

We observe that the straightforward projection introduces errors as demonstrated in Fig.~\ref{fig:drifts}.
Let $\Q$ be the projection matrix with rows ${\q_1, ..., \q_k}$, the projection is defined as $\pi(\th) = \Q \cdot \th$.
We model the projection of test LoRAs into the training embedding as a simple variation of domain adaptation~\cite{pan2008transfer,fernando2013unsupervised,hendy2024tl} and correct the errors with an affine transformation.

Let $\th'_i$ be a LoRA weight vector of an artist $i$ in the test LoRA set that is unseen by the PCA computation, and let $\{\th_{i,j}\!\mid\!i = 1, ..., n, \, j = 1, ..., m_{\text{train}}\}$ be the data fed to PCA.
We compute the average LoRA weights per artist as $\bar{\th}'_{i} = \sum_j \th'_{i,j}/m_{\text{test}}$, $\bar{\th}_{i} = \sum_j \th_{i,j}/m_{\text{train}}$.
We solve for the affine transformation with $\s = \{s_1, ..., s_k \}$ and $\t = \{t_1, ..., t_k \}$
\begin{equation}
	\begin{gathered}
	    \min_{s_k,t_k} \, \frac{1}{2} \sum_i \| s_k \cdot \pi_k(\bar{\th}'_i) + t_k - \pi_k(\bar{\th}_i) \|^2,\\
	    S_k^1 = \sum_i \pi_k(\bar{\th}_i) \cdot \pi_k(\bar{\th}'_i), \, S_k^2 = \sum_i \pi_k(\bar{\th}'_i)^2, \\
	    S^3_k =  \sum_i \pi_k(\bar{\th}_i), \, S^4_k = \sum_i \pi_k(\bar{\th}'_i)\\
	    s_k = \frac{S^1_k - (S^3_k \cdot S^4_k)/n}{S^2_k - (S^4_k)^2/n}, \, t_k =  \frac{S^3_k - s_k \cdot S^4_k}{n}.
	\end{gathered}
\end{equation}
With the $\s$, $\t$, we define a new projection $\pi(\th', \s, \t) = \text{diag}(\s) \, \Q \cdot \th + \t$ for unseen input LoRAs.

The remaining issue is the unknown true clustering of test LoRAs, preventing the construction of $\bar{\th}'_{i}$.
We address this by introducing a calibration set of LoRAs with known labels ($m_{\text{cali}}$ per artist), fine-tuned under the same configuration as the test LoRAs.

\section{Retrieval Results}

\begin{figure*}[htb!]
    \centering
    \includegraphics[width=\linewidth,height=\textheight,keepaspectratio]{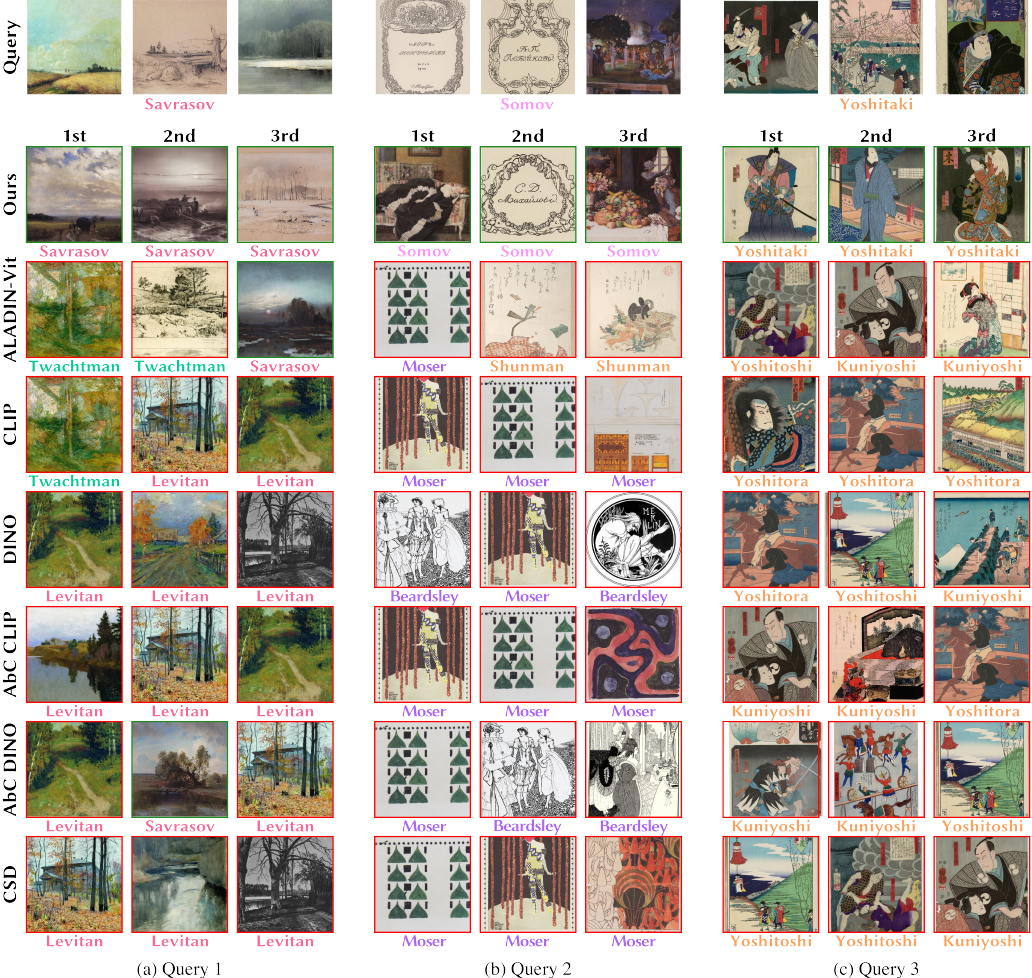}
  \caption{
    Example retrieval results.
    Artists' names are colored by the corresponding genres:
        \textcolor[HTML]{8036E0}{art nouveau},
        \textcolor[HTML]{FF6692}{realism},
        \textcolor[HTML]{00AA79}{impressionism},
        \textcolor[HTML]{FF97FF}{symbolism}, 
        and \textcolor[HTML]{E57C35}{ukiyo-e}.
    Representative images of the retrieved results are framed in green for correct answers and red for incorrect ones.
  }
  \label{fig:retrieval_results}
\end{figure*}

We present in Fig.~\ref{fig:retrieval_results} the top 3 retrieval results for the comparisons and our method in the \texttt{GenR} setting (Sec.~\ref{sec:retrieval}), where query images for comparisons are generated using descriptions of real images.
While this setting presents a mild challenge for the comparisons, the generated images still remain similar to the actual stylized images (examples in Fig.~\ref{fig:example}).
While we show all three images used as query, due to the space limit, we only show one representative image out of ten for one database sample.

Despite these challenging queries, our approach consistently outperforms the comparisons, achieving near-perfect results. 
\textcolor[HTML]{FF6692}{Realism} paintings emphasize accurate depictions of everyday life, featuring precise details, natural color palettes, realistic lighting, and lifelike subjects such as people, landscapes, and ordinary scenes (Fig.~\ref{fig:retrieval_results}a).
Although most comparison methods retrieve results within the same genre (or the related \textcolor[HTML]{00AA79}{impressionism} genre) and similar landscape content, they consistently fail to return correct results.
Among them, ALADIN-ViT~\cite{ruta2021aladin} and AbC DINO~\cite{wang2023evaluating} achieve only partial correctness.
Query 2 (Fig.~\ref{fig:retrieval_results}b) is an interesting example consisting of Somov's ornamental title pages, which differ significantly from his more well-known oil paintings.
While our approach retrieves the correct answers, all comparison methods are confused by \textcolor[HTML]{8036E0}{art nouveau} and \textcolor[HTML]{E57C35}{ukiyo-e} artists due to their printed artwork.
The final example query demonstrates that our approach can distinguish subtle differences between teachers (Utagawa Kuniyoshi) and apprentices (Utagawa Yoshitaki) (Fig.~\ref{fig:retrieval_results}c). 
Similar to Query 1 (Fig.~\ref{fig:retrieval_results}a), all comparison methods return \textcolor[HTML]{E57C35}{ukiyo-e} results, but the retrieved artists are incorrect.

\begin{figure}[t]
    \centering
  \includegraphics[width=\linewidth]{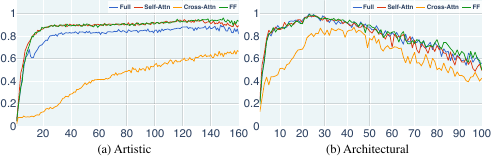}
  \caption{
  Validation set clustering performance varies across sub-networks --- feedforward network (FF), self-attention (Self-Attn), and cross-attention (Cross-Attn) --- with similar trends on Artistic (a) and Architectural (b).
  }
  \label{fig:ari_curves}
\end{figure}

\begin{figure}[t]
  \includegraphics[width=\linewidth]{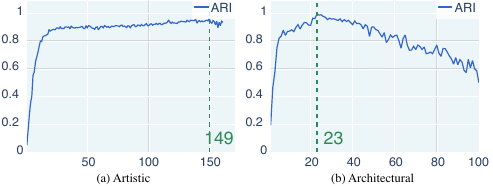}
  \caption{
  Clustering performance as a function of the number of PCs on (a) Artistic and (b) Architectural databases.
    We determine the optimal number of PCs (indicated by green dashed lines) in Sec.~\ref{sec:ablations} using the validation set.
  }
  \label{fig:full_curves}
\end{figure}

\section{Ablation Details}

\paragraph{PCA}
We visualize the clustering performance curves of the feedforward network on the validation set.
As shown in Fig.~\ref{fig:full_curves}, performance rises sharply within the range $[1, 10]$ (Artistic) and $[1, 5]$ (Architectural), peaks at an optimum, and then gradually declines.

\paragraph{Sub-networks}
In Fig.~\ref{fig:ari_curves}, we plot the clustering performance of the full U-Net and its three sub-networks with different dimensionalities: feedforward network ($0.47d$), self-attention ($0.27d$), and cross-attention ($0.27d$).
By default, we select feedforward network weights as our weight vectors in evaluations.

\begin{figure*}[t]
  \includegraphics[width=\linewidth]{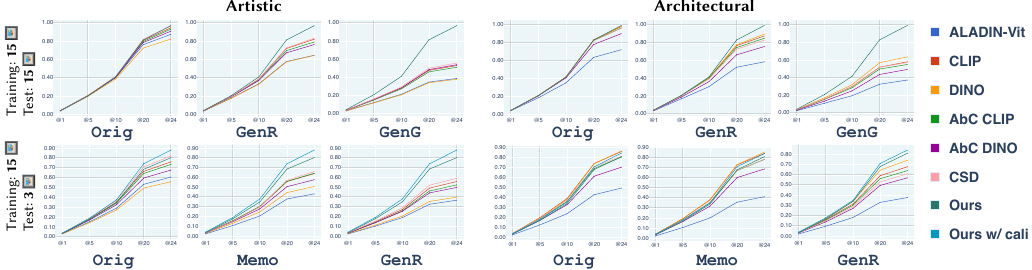}
  \caption{
  Recall@k curves corresponding to our recall results shown in the main paper.
  Performance rankings across methods are generally consistent across different Recall@k values.
  }
  \label{fig:recall_k}
\end{figure*}

\begin{figure}[t]
  \includegraphics[width=\linewidth]{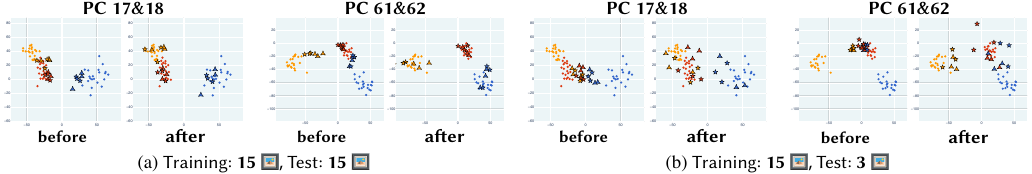}
  \caption{
  Our calibration method corrects for the drift between the training LoRA embeddings ($\mathbf{\bullet}$) and the test LoRA embeddings ($\bigstar$) by using the hold-out calibration set ($\blacktriangle$) to calibrate.
  Before the calibration, the test LoRA embeddings are misaligned from the train LoRAs (top row). 
  After calibration, the distributions are closer together (bottom row). 
  For matching fine-tuning set sizes, simple scaling (equivalent to cosine distance) effectively corrects errors (a).
  Our method generalizes reasonably to test LoRAs trained under different configurations (bottom right, b).
  }
  \label{fig:cali}
\end{figure}

\paragraph{Calibration}
We provide a visual example showing that our calibration in Sec.~\ref{sec:ablations} produces better-aligned projections in both types of test LoRAs (Fig.~\ref{fig:cali}).

\section{Additional Experiments}

\paragraph{Data domains}
We conduct two small-scale validation experiments on additional data domains: content (ImageNet~\cite{deng2009imagenet}: 7 categories) and identity (VGGFace2~\cite{cao2018vggface2}: 7 individuals).  
We apply the same preprocessing protocol and train 20 SD1.5 LoRAs per category or individual.  
To remove hints from auto-captions, we replace category names with ``object'' and substitute gendered words with gender-neutral ones.  
Both experiments yield well-separated clusters.  
Interestingly, in the ImageNet experiment, the closest cluster pair under the first two principal components (PC 1 and 2) is car and truck, among the seven categories: airship, car, penguin, plane, ship, snow leopard, truck (Fig.).

\paragraph{Rank}
We further study LoRAs of different ranks using a subset of 12 artists from our SDXL experiment, with 30 LoRAs per artist at rank 16 (our default) and 10 LoRAs per artist at rank 8 and 32.  
Let $\theta$ and $\phi$ denote the LoRA vectors of rank 16 and 8 (or 32), respectively, and let $P$ and $Q$ be the PCA projection matrices fit to $\theta$ and $\phi$.  

When projecting both $\theta$ and $\phi$ using $P$, we truncate $\theta$ to match the dimensionality of $\phi$.  
The projected vectors $P\phi$ concentrate near the PCA origin, with magnitudes about $1/15$ that of $P\theta$, leading to partially overlapping clusters.  
Nevertheless, the outermost clusters remain distinguishable, indicating preserved structure despite strong compression.  

When projecting $\theta$ and $\phi$ with their respective PCA matrices $P$ and $Q$, we find $P\theta \approx Q\phi$ up to sign flips (due to PCA’s arbitrary orientation).  
Quantitatively, repeating our clustering experiment (Sec.~\ref{sec:clustering}) with $\phi$ (rank 8 and 32) as the test set yields NMI = 99.7\% and 96.2\% respectively (100\% being perfect).  
This alignment across ranks is significantly stronger than that between LoRA weights and image-based features:  

\begin{table}[h]
\centering
\scriptsize
\setlength{\tabcolsep}{2pt} %
\begin{tabular*}{\columnwidth}{@{\extracolsep{\fill}}lcccccccc}
\toprule
Metric & \shortstack{LoRA\\(8 $\rightarrow$ 16)} & \shortstack{LoRA\\(32 $\rightarrow$ 16)} & ALADIN & CLIP & DINO & AbC-C & AbC-D & CSD \\
\midrule
ARI $\uparrow$ & 98.8\% & 89.9\% & 3.8\% & 9.0\% & 9.5\% & 11.2\% & 7.3\% & 11.1\% \\
NMI $\uparrow$ & 99.7\% & 96.2\% & 20.6\% & 33.0\% & 30.9\% & 39.2\% & 31.4\% & 36.9\% \\
\bottomrule
\end{tabular*}
\end{table}

These results validate that structural similarity across LoRAs of different ranks is not incidental: their embeddings preserve a consistent structure up to orthogonal transformations.  
In practice, this suggests a simple alignment strategy: (1) bin LoRAs by rank, (2) construct PCA embedding per bin, and (3) align embeddings by adjusting signs.

\section{Detailed Statistics}

Retrieval metric statistics (Fig.~\ref{fig:retrieval_stats}, Sec.~\ref{sec:retrieval}) and results for two additional database and query image settings (\texttt{Orig+GenR}, \texttt{GenR+Orig}) are presented in Table~\ref{tab:full_retrieval}.
Note that these additional settings are less favorable to the baselines compared to \texttt{GenR}.
The detailed measures for embedding similarity (inset in Sec.~\ref{sec:embed_sim}) are shown in Table~\ref{tab:sim}.

Due to space constraints, we present only Recall@10 in the main paper.
Performance rankings across methods are generally consistent across different Recall@k values.
Here, we provide all detailed Recall@k curves in Fig.~\ref{fig:recall_k}.

\begin{table*}[htbp]
\small
\caption{
    Retrieval performance comparison (mean $\pm$ std) between image-based features and our approach across different scenarios.
    Higher is better for both metrics.
}
\centering

\begin{subtable}[t]{\textwidth}
\caption{Artistic fine-tuning set of 15 images.}

\resizebox{\textwidth}{!}{
\begin{tabular*}{1.8\textwidth}{@{\extracolsep{\fill}}l|ccccc|ccccc}
\hlineB{3}
\multirow{2}{*}{Methods} & \multicolumn{5}{c|}{mAP $\uparrow$ (\%)} & \multicolumn{5}{c}{Recall@10 $\uparrow$ (\%)} \\
 & \texttt{Orig} & \texttt{Orig+GenR} & \texttt{GenR+Orig} & \texttt{GenR} & \texttt{GenG} & \texttt{Orig} & \texttt{Orig+GenR} & \texttt{GenR+Orig} & \texttt{GenR} & \texttt{GenG} \\
\hlineB{3}
ALADIN-Vit&$96.3 \pm 9.4$&$69.6 \pm 32.6$&$75.1 \pm 23.9$&$82.8 \pm 19.9$&$59.6 \pm 29.4$&$40.1 \pm 5.0$&$27.0 \pm 15.1$&$29.3 \pm 11.9$&$33.4 \pm 10.1$&$21.0 \pm 13.5$\\
CLIP&$98.8 \pm 4.8$&$91.4 \pm 18.1$&$91.6 \pm 13.6$&$92.8 \pm 14.0$&$73.2 \pm 25.1$&$41.2 \pm 2.4$&$37.8 \pm 8.7$&$38.0 \pm 6.9$&$38.5 \pm 7.1$&$28.2 \pm 12.6$\\
DINO&$93.8 \pm 13.6$&$82.2 \pm 25.5$&$82.3 \pm 21.9$&$82.2 \pm 22.1$&$58.0 \pm 29.8$&$38.8 \pm 7.1$&$33.1 \pm 12.5$&$33.1 \pm 10.8$&$32.9 \pm 11.1$&$20.5 \pm 13.4$\\
AbC CLIP&$98.0 \pm 7.5$&$84.3 \pm 25.3$&$87.7 \pm 17.8$&$91.2 \pm 15.3$&$71.1 \pm 27.1$&$40.8 \pm 4.0$&$34.3 \pm 12.0$&$36.1 \pm 8.8$&$37.7 \pm 7.7$&$27.1 \pm 13.3$\\
AbC DINO&$97.5 \pm 8.2$&$86.0 \pm 23.4$&$85.7 \pm 20.1$&$90.0 \pm 16.4$&$73.5 \pm 26.7$&$40.6 \pm 4.3$&$34.9 \pm 11.4$&$34.7 \pm 10.1$&$37.0 \pm 8.3$&$28.2 \pm 13.3$\\
CSD&$\mathbf{99.4 \pm 3.0}$&$89.0 \pm 21.9$&$90.8 \pm 17.0$&$93.3 \pm 13.9$&$75.0 \pm 25.5$&$\mathbf{41.5 \pm 1.5}$&$36.5 \pm 10.5$&$37.4 \pm 8.4$&$38.7 \pm 6.9$&$29.1 \pm 12.8$\\
\hline
Ours&$98.4 \pm 9.4$&$\mathbf{98.4 \pm 9.4}$&$\mathbf{98.4 \pm 9.4}$&$\mathbf{98.4 \pm 9.4}$&$\mathbf{98.4 \pm 9.4}$&$41.0 \pm 4.3$&$\mathbf{41.0 \pm 4.3}$&$\mathbf{41.0 \pm 4.3}$&$\mathbf{41.0 \pm 4.3}$&$\mathbf{41.0 \pm 4.3}$\\
\hlineB{3}
\end{tabular*}
}
\end{subtable}

\vspace{1em}

\begin{subtable}[t]{\textwidth}

\caption{Architectural fine-tuning set of 15 images.}
\resizebox{\textwidth}{!}{
\begin{tabular*}{1.8\textwidth}{@{\extracolsep{\fill}}l|ccccc|ccccc}
\hlineB{3}
\multirow{2}{*}{Methods} & \multicolumn{5}{c|}{mAP $\uparrow$ (\%)} & \multicolumn{5}{c}{Recall@10 $\uparrow$ (\%)} \\
 & \texttt{Orig} & \texttt{Orig+GenR} & \texttt{GenR+Orig} & \texttt{GenR} & \texttt{GenG} & \texttt{Orig} & \texttt{Orig+GenR} & \texttt{GenR+Orig} & \texttt{GenR} & \texttt{GenG} \\
\hlineB{3}
ALADIN-Vit&$85.9 \pm 17.5$&$62.3 \pm 27.0$&$68.9 \pm 25.8$&$77.9 \pm 18.8$&$54.6 \pm 25.3$&$35.2 \pm 8.9$&$23.5 \pm 13.1$&$27.1 \pm 12.5$&$30.7 \pm 9.6$&$19.5 \pm 11.2$\\
CLIP&$99.9 \pm 1.0$&$95.6 \pm 14.7$&$97.1 \pm 6.4$&$96.6 \pm 7.3$&$75.4 \pm 25.5$&$41.6 \pm 0.7$&$39.8 \pm 6.8$&$40.5 \pm 3.4$&$40.5 \pm 3.6$&$29.9 \pm 12.6$\\
DINO&$99.9 \pm 0.4$&$95.6 \pm 13.1$&$98.3 \pm 4.2$&$97.2 \pm 7.1$&$81.4 \pm 22.5$&$41.7 \pm 0.0$&$39.8 \pm 6.1$&$40.9 \pm 2.7$&$40.6 \pm 3.5$&$32.0 \pm 11.8$\\
AbC CLIP&$99.8 \pm 0.7$&$92.4 \pm 18.6$&$96.4 \pm 6.2$&$95.7 \pm 7.2$&$74.2 \pm 25.0$&$41.6 \pm 0.4$&$38.2 \pm 8.7$&$40.2 \pm 3.5$&$40.2 \pm 3.6$&$28.6 \pm 12.4$\\
AbC DINO&$97.7 \pm 4.8$&$88.1 \pm 18.6$&$91.0 \pm 11.8$&$88.8 \pm 14.5$&$66.1 \pm 25.0$&$40.7 \pm 2.8$&$35.8 \pm 10.0$&$37.8 \pm 5.7$&$36.3 \pm 7.4$&$25.0 \pm 12.5$\\
CSD&$99.6 \pm 1.0$&$90.9 \pm 16.9$&$92.5 \pm 12.7$&$94.7 \pm 8.5$&$75.7 \pm 24.5$&$41.7 \pm 0.0$&$37.7 \pm 8.2$&$38.3 \pm 6.5$&$39.3 \pm 4.7$&$29.8 \pm 11.9$\\
\hline
Ours&\multicolumn{5}{c|}{$\mathbf{99.9 \pm 1.2}$}&\multicolumn{5}{c}{$\mathbf{41.6 \pm 0.7}$}\\
\hlineB{3}
\end{tabular*}
}
\end{subtable}

\vspace{1em}

\begin{subtable}[t]{\textwidth}

\caption{Artistic fine-tuning set of 3 images.}
\resizebox{\textwidth}{!}{
\begin{tabular*}{1.02\textwidth}{@{\extracolsep{\fill}}l|ccc|ccc}
\hlineB{3}
\multirow{2}{*}{Methods} & \multicolumn{3}{c|}{mAP $\uparrow$ (\%)} & \multicolumn{3}{c}{Recall@10 $\uparrow$ (\%)} \\
 & \texttt{Orig} & \texttt{Memo} & \texttt{GenR} & \texttt{Orig} & \texttt{Memo} & \texttt{GenR} \\
\hlineB{3}
ALADIN-Vit&$73.6 \pm 31.8$&$53.8 \pm 38.3$&$51.0 \pm 34.6$&$29.2 \pm 14.7$&$20.6 \pm 16.8$&$18.8 \pm 15.2$\\
CLIP&$86.8 \pm 23.0$&$74.8 \pm 31.4$&$69.5 \pm 30.7$&$35.5 \pm 10.9$&$30.0 \pm 14.5$&$27.2 \pm 14.2$\\
DINO&$70.2 \pm 32.1$&$63.8 \pm 34.6$&$55.4 \pm 33.9$&$27.4 \pm 14.9$&$24.7 \pm 15.7$&$20.5 \pm 15.1$\\
AbC CLIP&$84.4 \pm 25.2$&$75.6 \pm 31.0$&$66.4 \pm 31.9$&$34.4 \pm 12.0$&$30.4 \pm 14.2$&$25.8 \pm 14.5$\\
AbC DINO&$81.6 \pm 26.5$&$71.3 \pm 33.6$&$66.0 \pm 32.3$&$32.7 \pm 12.8$&$27.8 \pm 15.6$&$25.3 \pm 14.9$\\
CSD&$90.5 \pm 19.6$&$74.3 \pm 34.0$&$73.1 \pm 30.2$&$37.2 \pm 9.6$&$29.9 \pm 15.3$&$28.9 \pm 14.2$\\
\hline
Ours&\multicolumn{3}{c|}{$85.3 \pm 28.3$}&\multicolumn{3}{c}{$35.0 \pm 13.0$}\\
Ours w/ cali&\multicolumn{3}{c|}{$\mathbf{91.1 \pm 23.2}$}&\multicolumn{3}{c}{$\mathbf{38.0 \pm 10.0}$}\\
\hlineB{3}
\end{tabular*}
}
\end{subtable}

\vspace{1em}

\begin{subtable}[t]{\textwidth}

\caption{Architectural fine-tuning set of 3 images.}
\resizebox{\textwidth}{!}{
\begin{tabular*}{1.02\textwidth}{@{\extracolsep{\fill}}l|ccc|ccc}
\hlineB{3}
\multirow{2}{*}{Methods} & \multicolumn{3}{c|}{mAP $\uparrow$ (\%)} & \multicolumn{3}{c}{Recall@10 $\uparrow$ (\%)} \\
 & \texttt{Orig} & \texttt{Memo} & \texttt{GenR} & \texttt{Orig} & \texttt{Memo} & \texttt{GenR} \\
\hlineB{3}

ALADIN-Vit&$62.1 \pm 30.3$&$54.4 \pm 30.0$&$50.2 \pm 31.8$&$23.7 \pm 14.3$&$20.1 \pm 13.7$&$18.3 \pm 14.0$\\
CLIP&$\mathbf{94.1 \pm 13.6}$&$92.4 \pm 16.6$&$78.5 \pm 29.4$&$\mathbf{38.9 \pm 7.0}$&$\mathbf{38.1 \pm 8.3}$&$31.4 \pm 13.8$\\
DINO&$92.3 \pm 18.1$&$\mathbf{92.5 \pm 17.4}$&$84.0 \pm 26.5$&$38.1 \pm 8.8$&$37.9 \pm 9.0$&$34.1 \pm 12.8$\\
AbC CLIP&$88.6 \pm 20.5$&$86.6 \pm 22.0$&$76.5 \pm 29.8$&$36.0 \pm 10.4$&$35.1 \pm 11.0$&$30.3 \pm 14.2$\\
AbC DINO&$81.1 \pm 27.6$&$79.2 \pm 29.0$&$69.2 \pm 32.7$&$32.7 \pm 13.0$&$31.8 \pm 13.5$&$27.2 \pm 15.2$\\
CSD&$90.7 \pm 17.1$&$86.7 \pm 20.8$&$72.9 \pm 29.7$&$37.3 \pm 8.6$&$35.6 \pm 10.2$&$28.7 \pm 13.9$\\
\hline
Ours&$82.4 \pm 34.6$&$82.4 \pm 34.6$&$82.4 \pm 34.6$&$33.9 \pm 15.0$&$33.9 \pm 15.0$&$33.9 \pm 15.0$\\
Ours w/ cali&$86.5 \pm 31.3$&$86.0 \pm 31.0$&$\mathbf{86.0 \pm 31.0}$&$35.4 \pm 14.0$&$35.0 \pm 14.0$&$\mathbf{35.0 \pm 14.0}$\\
\hlineB{3}
\end{tabular*}
}
\end{subtable}

\label{tab:full_retrieval}
\end{table*}

\end{document}